\documentclass[letterpaper, 10 pt, conference]{ieeeconf}  % Comment this line out if you need a4paper

\IEEEoverridecommandlockouts                              % This command is only needed if 
\usepackage{graphics} % for pdf, bitmapped graphics files
\usepackage{epsfig} % for postscript graphics files
\usepackage{times} % assumes new font selection scheme installed
\usepackage{amsmath} % assumes amsmath package installed
\usepackage{amssymb}  % assumes amsmath package installed
\usepackage{amsfonts}
\usepackage{multirow}
\usepackage{breqn}
\usepackage{xcolor}
\usepackage{algorithm}
\usepackage{algorithmic}
\usepackage{mathrsfs}
\usepackage{booktabs} % For better table lines

\title{\LARGE \bf
Relevance for Human Robot Collaboration
}

\author{Xiaotong Zhang, Dingcheng Huang and Kamal Youcef-Toumi% <-this % stops a space
\thanks{All authors are with the Mechatronics Research Laboratory, Massachusetts Institute of Technology, Cambridge, MA, 02139, USA. Xiaotong Zhang and Dingcheng Huang contributed equally.
{\tt\small \{kevxt, dean1231, youcef\}@mit.edu }}% <-this % stops a space
\thanks{This research was made possible by the support and partnership
of King Abudlaziz City for Science and Technology
(KACST) through the Center for Complex Engineering
Systems at Massachusetts Institute of Technology (MIT) and
KACST.}
}

\begin{document}

\maketitle
\thispagestyle{empty}
\pagestyle{empty}

%%%%%%%%%%%%%%%%%%%%%%%%%%%%%%%%%%%%%%%%%%%%%%%%%%%%%%%%%%%%%%%%%%%%%%%%%%%%%%%%
\begin{abstract}
%Effective human-robot collaboration (HRC) requires the robots to possess human-like intelligence. 
Inspired by the human ability to selectively focus on relevant information, this paper introduces relevance, a novel dimensionality reduction process for human-robot collaboration (HRC). Our approach incorporates a continuously operating perception module, evaluates cue sufficiency within the scene, and applies a flexible
formulation and computation framework. To accurately and efficiently quantify relevance, we developed an event-based framework that maintains a continuous perception of the scene and selectively triggers relevance determination. Within this framework, we developed a probabilistic methodology, which considers various factors and is built on a novel structured scene representation. 
%Then, the framework identifies and classifies relevant elements based on features in the scene, enabling robots to operate more intelligently and efficiently in real-world environments. By incorporating event-based relevance determination and probability-based quantification, our framework significantly enhances human-robot interaction, task planning, decision-making, and perception performance. 
Simulation results demonstrate that the relevance framework and methodology accurately predict the relevance of a general HRC setup, achieving a precision of 0.99, a recall of 0.94, an F1 score of 0.96, and an object ratio of 0.94. Relevance can be broadly applied to several areas in HRC to accurately improve task planning time by 79.56\% compared with pure planning for a cereal task, reduce perception latency by up to 26.53\% for an object detector, improve HRC safety by up to 13.50\% and reduce the number of inquiries for HRC by 80.84\%. A real-world demonstration showcases the relevance framework's ability to intelligently and seamlessly assist humans in everyday tasks.
\end{abstract}

%%%%%%%%%%%%%%%%%%%%%%%%%%%%%%%%%%%%%%%%%%%%%%%%%%%%%%%%%%%%%%%%%%%%%%%%%%%%%%%%

\section{Introduction}

Robots and automated systems are increasingly integrated into modern society, and the human world, which requires them to develop human-like intelligence and cognitive abilities \cite{zhang2022systematic, zhang2022magnetohydrodynamic}. One example of such abilities is the human capability to selectively focus on relevant physical elements in the environment and relevant abstract concepts or information in the input, guided by context, objectives, historical experiences, and reasoning. This cognitive function is closely associated with the reticular activating system (RAS), a network of neurons in the brain responsible for filtering sensory information, capturing relevant details, and suppressing irrelevant stimuli to optimize cognitive processing and decision-making \cite{moruzzi1949brain}. Various studies have demonstrated the aforementioned human ability. \cite{mack2003inattentional, stewart2020review}.

Empowering robots with this human-like cognitive capability offers two significant benefits. First, robots interacting closely with human beings will achieve a scene-understanding similar to that of humans. This results in more natural and effective behavior and interaction \cite{zhang2024relevancedrivendecisionmakingsafer}. Second, by identifying the most relevant elements in the scene, the robot can effectively focus its limited computation power on the relevant elements to reduce the computation requirements, increase the computation speed, and even increase the safety of the robots in HRC tasks \cite{zhang2024does}.

In this paper, we define and introduce a novel concept and approach for dimensionality reduction, termed `relevance.' %Relevance is a process of dimension reduction that involves continuously perceiving the scene, identifying and detecting elements in the scene, classifying elements into a hierarchical representation, e.g., classes, and utilizing cues to selectively reduce the dimension of inputs. If the cues are insufficient, relevance further involves gathering additional cues iteratively and reapplying dimension reduction once cues are sufficient.
Relevance is
defined as a
dimensionality
reduction process
that continuously
perceives the
scene, identifies
and detects its
elements, and
organizes them
into a hierarchical
representation,
such as attribute
classes. This
process utilizes
contextual cues
to selectively
reduce input
dimensionality.
When cues are
insufficient,
relevance-driven
processing
iteratively gathers
additional cues
and reapplies
dimensionality
reduction until
sufficient
information is
obtained.

Previous works in this area to identify prominent features have primarily focused on visual saliency and attention mechanisms. Visual saliency aims to identify the most distinctive regions within an image based on the image features, such as color, intensity, and orientation \cite{itti1998model}. As a result, the salient regions detected are mostly visually conspicuous regions. However, visual saliency does not consider context information, which is critical in the application of HRC. Attention is a type of mechanism to build a mapping from the embedded input data to importance weights through a structured learnable function. The attention mechanism acts as an add-on module inside neural network models to consider the weights of each component to improve performance. Attention has been widely adopted and implemented in various applications, such as end-to-end frameworks to generate actions from sensory inputs \cite{zhu2023viola}, various sub-functions in the robotic pipelines \cite{carion2020end, kedia2024interact}, etc.

Compared with saliency and attention, relevance is a comprehensive framework and concept that dramatically extends the scope and functionalities of these works. Our relevance, in its current form, already considers context information, such as human objectives, tasks, environment information, human preferences, etc. These factors are essential to enable proactive robotics assistance. Moreover, relevance is not only a module to generate the importance weight of each object in the input but also a novel framework that integrates a continuously running multi-modality perception module, a novel hierarchical scene representation, consideration of cue sufficiency in the input, etc. Those functions of relevance, which are not considered in saliency and attention, are essential for accurate, reliable, and seamless HRC. Last but not least, our relevance features a flexible formulation and computational principle not limited to the structure of neural networks. In this paper, as an illustration. we develop a probabilistic model that leverages an AI toolkit with a large language model (LLM), human preferences, low-level attributes, and constraints to generate the necessary information. This structure and formulation enable anticipatory capabilities, interactions among different information sources, a far more complex reasoning process, etc, which finally aims for artificial general intelligence (AGI). The demonstration in this paper shows, with relevance, that the robot can generate optimal reactions and seamless assistance to humans without transfer learning.

%Previous works in this area to identify prominent features have primarily focused on visual saliency and attention mechanisms. However, they often overlook the broader context necessary for long-term task planning and proactive assistance. Additionally, attention mechanisms are typically constrained to short-term simple motion generation and lack interpretability, which makes them unsuitable for complex problems requiring comprehensive reasoning. These limitations highlight the need for a novel approach designed for context-aware, efficient, and accurate scene understanding, reasoning about the interconnection among humans, dynamic environments, context, etc. 

%In relevance, our approach to scene understanding starts by identifying all elements in the scene, which are then grouped into classes. Our approach identifies the classes of contextual elements pertinent to what is or may happen in the scene. Our approach helps identify all classes with supplementary elements. Such classes/elements are not core to the scene with a positive (enhancing), negative (distracting), or neutral (optional) effect. Finally, the approach identifies the core or essential elements, focusing on indispensable components like main characters, primary actions, and key elements for what is taking place. This approach ensures a structured, coherent, systematic, and efficient scene understanding that is updated in real time.

To quantify relevance, we propose and develop a novel framework. The event-based framework contains modules of perception, triggers, relevance determination, and decision-making. The perception module is continuously running and perceives cues in the environment based on multi-modality sensors and processing. The event-based mechanism initializes the relevance determination only under certain circumstances, resulting in a dramatic increase in computation speed. %In this framework, the perception algorithms are running continuously and monitoring the features in the scene. %Some examples of the features include features from vision (element detection and characterization, and motion tracking), audio (conversation and background noise), context (objective, tasks, preferences), etc. 
%Once some criteria are met based on the features, a trigger is fired, and the relevance determination is initialized. Then, the relevant classes and elements are quantified and determined, followed by decision-making to assist and facilitate human beings faster, more naturally, and safer. 
In the relevance determination module, flexible formulations and computation principles can be applied. As an illustration, we developed a unified probabilistic methodology based on a novel scene representation. This methodology accurately quantifies the relevance of the scene, which dramatically improves the performance of HRC. Finally, a decision making module generates actions based on the results of relevance determination.

%In this scene representation, the detected elements are grouped into classes based on certain classification criteria, including functional criteria and attribute criteria. We define the relevance of a class or an element as the probability they are relevant given the objective, which can be quantified with a unified probability framework. Various factors considered can be modeled as a probability and integrated into the quantification. 

% \textcolor{orange}{This paragraph needs to discuss the effectiveness we achieved in this paper. Update later. Several effects of relevance: 1) lower perception latency because of the framework. 2) natural and fewer inquiries because of relevance. 3) TAMP??}

To summarize, the contributions of this paper are four-fold: 1. We proposed a new concept, dimensionality reduction approach, and framework called relevance to enable fast, accurate, seamless, and proactive HRC. 2. We developed an event-based framework for faster relevance determination. 3. We developed a probabilistic methodology for relevance quantification and determination based on a novel hierarchical scene representation. 4. We demonstrated the effectiveness of our framework and methodology using both simulation evaluation and real-world demonstration.

\section{Related Works}
Two areas of research, i.e., visual saliency and attention, share similar goals to our work but with much narrower scope and functionalities. We also include a review of related HRC literature. To the best of our knowledge, this work is unique in HRC that it empowers the robots with the cognitive capability to focus on selective components in the input by developing a novel framework and methodology with unique advantages, which are essential for HRC.

\subsection{Visual Saliency}
Visual saliency seeks to identify and localize the most distinctive regions or objects within an image based on visual conspicuity without considering the context. Traditional methods are developed based on the features at the pixel level, including colors, intensity, orientation, etc \cite{itti1998model, meger2008curious, chung2002new}. Deep learning-based methods can also be developed and trained with the help of visual saliency datasets that are annotated either manually with a mouse click or automatically with eye gaze tracker \cite{li2016visual,kuen2016recurrent,pan2017salgan,liu2021visual}. However, visual saliency mainly considers visual conspicuity, neglecting important considerations, such as context, for proactive assistance in HRC. 

\subsection{Attention}
Attention is another mechanism that selectively focuses on relevant parts of the input data but with limited scope and functionalities compared to relevance. In a nutshell, attention is structured with self-attention, cross-attention, multi-layer perception (MLP), etc., to formulate and optimize weights for the input features \cite{vaswani2017attention}. However, attention in robotics is mainly applied to end-to-end algorithms for better completion of the short-term action, e.g., pick and place \cite{zhu2023viola,wang2022transformer,devin2018deep}. Moreover, attention has several challenging limitations, such as limited applicability to end-to-end methods only, limited interpretability, and myopia. %It is trained and optimized along with other parameters in the neural networks to optimize the values of the objective functions. Another way to structure attention is without self-attention or cross-attention but to augment some components in the neural networks with weights explicitly, and the weights are to be optimized along with other parameters according to the objective functions through backpropagation. 
% Relevance detects and identifies the relevant classes or elements for the long-term objective and potential tasks based on a structured methodology to consider various factors, including human factors (motion, goals, objectives, preference), task models (task structure, task prediction, object applicability), etc. 
Relevance not only resembles the cognitive capability to reduce the dimension of original input but also alleviates the limitations of attention by developing a novel event-based framework and distinctive methodologies, such as a flexible formulation, continuously running perception, hierarchical scene representation, multi-level relevance determination, etc. Thus, our relevance can integrate the information and cues from an AI toolkit with LLMs, human preferences, human objectives, environmental constraints, etc., to enable accurate, seamless, and proactive assistance in HRC.

\subsection{Human Robot Collaboration}
There have been significant advancements in the fields related to HRC. We classify those works into two categories. The first category of works focuses on one specific important function for HRC, such as object detection \cite{zhang2024does}, semantic segmentation \cite{munasinghe2022covered}, human intention prediction \cite{9428016,BUERKLE2021102137,9437772,hernandezcruz2024bayesianintentionenhancedhuman}, human motion prediction \cite{kothari2023enhanced}, human-aware task planning \cite{9345470}, etc. Those works, though important, have fundamentally different purposes from relevance. They are not for dimensionality reduction. The results of those modules can be applied for relevance quantification, and relevance can, vice versa, benefit these modules. The second category of work aims to develop a framework that directly generates actions from sensory inputs. Those works include learning from demonstration \cite{8403899}, reinforcement learning \cite{el2020towards}, etc. To incorporate the selective processing idea into these functions, some work leverage attention mechanisms if the model is neural network-based \cite{zhang2022graph}. However, because of the narrower scope of attention, as described in Section II.B, our relevance possesses more functions and is more advanced than attention. As shown in Sections V and VI, with relevance, robots can generate proactive, seamless, accurate, and structured reactions in an autonomous manner.

\section{Framework for Relevance}
\begin{figure*}[!t]
\begin{center}
\includegraphics[width=1.0\textwidth]{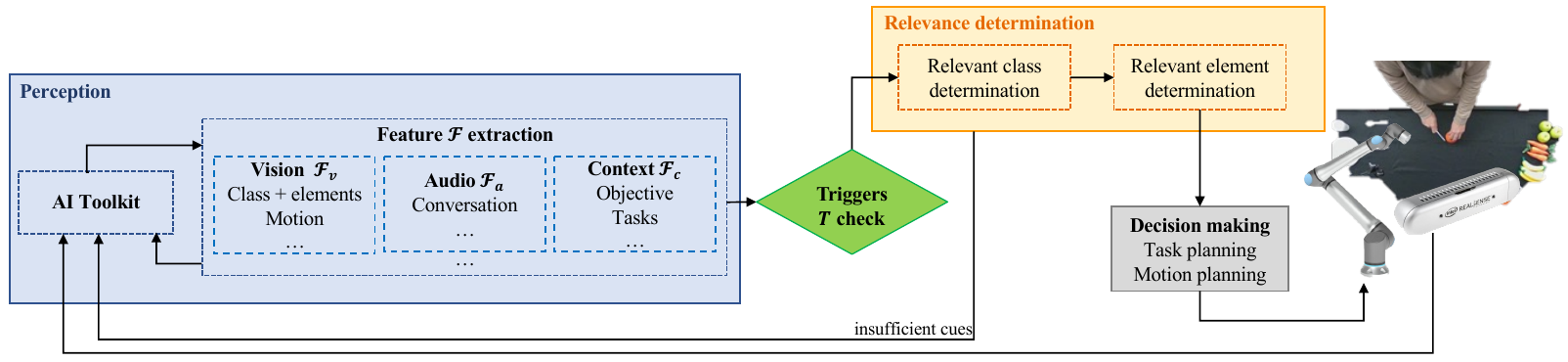}
\end{center}
\caption{Overview of the framework for relevance quantification and application of relevance for proactive human-robot collaboration. The framework consists of a continuously running perception module, a trigger check module to selectively initialize relevance determination, a two-level relevance determination methodology, and a decision-making module to generate natural and efficient human robot interaction.}
\label{fig: framework}
\end{figure*}

In this section, we introduce our novel event-based framework for relevance determination and robotic sensorimotor action generation, as shown in Fig. \ref{fig: framework}. The framework contains perception module, triggers $T$ check, relevance determination, and decision making.

% \subsection{Framework Overview}
% \textcolor{blue}{The overview of the framework is shown in Fig. \ref{fig: framework}.} On the right is a real-world setup that contains sensors, robots, other agents, a dynamic environment, etc. The perception module runs continuously mapping sensory inputs to the features of the scene $\mathcal{F}$. Once a trigger is fired because of some features, the relevance determination module is initialized to determine the relevant classes, relevant elements, and if the information is sufficient. If the information is insufficient, the algorithm will come back to the perception module for more information. Followed by relevance is a decision-making module utilizing the relevance for advanced and proactive action generation.

\subsection{Perception module}
The perception module continuously processes the sensor inputs (camera, microphone, etc.) with an AI toolkit and obtains the features $\mathcal{F}$ of the scene. Unlike other pipeline frameworks for robots that process the perception and decision-making in a serial manner \cite{kothari2023enhanced}, our developed and proposed framework perceives the scene in an uninterrupted manner to capture the fast and dynamic changes in the scene. Potential algorithms in the AI toolkit include You Only Look Once (YOLO) \cite{redmon2016you}, MMPose \cite{mmpose2020}, CLIP \cite{radford2021learningtransferablevisualmodels}, Visual Language Models (VLM) \cite{alayrac2022flamingo}, etc.
%Some potential algorithms can be leveraged here, including object detection, human pose estimation, human objective prediction, audio processing, natural language processing (NLP), large language models (LLM), visual language models, etc. 

\subsection{Triggers Check}
The triggers $T$ are to determine if a subprocess should be initialized for relevance determination based on the features $\mathcal{F}$ derived from the perception module. Some potential criteria for trigger activation include element changes and/or updated objectives or new tasks for the human. %Element addition corresponds to relevance reevaluation because the relevance of the new element is absent, and the newly added element could become relevant instead of the originally relevant elements. Removal of an element may also require rebalancing and readjustment of the relevance of each element to account for the compensation of the removed element. A new objective or new task will require the relevance to be reevaluated because the major factors that determine the utility and applicability of elements in the scene to the human have changed. 

\subsection{Relevance Determination}
One unique advantage of relevance is that it enables flexible formulations and computation principles. In this paper, as an illustration, our relevance determination methodology is a probability framework based on a new type of scene representation, i.e., classes and elements. The relevance determination methodology is a two-step methodology that determines the relevance classes and the relevance elements. At the same time, the methodology determines if sufficient information is available. With sufficient information, the framework will go through the methodology and determine relevance of the scene. With insufficient information, the algorithms will request or wait for the perception module to give out more information and cues. More details of our methodology for relevance determination will be discussed in Section IV. 

\subsection{Decision Making}
The output of the relevance determination module will be utilized in the decision-making to provide natural, faster, and safer actions for proactive HRC.

\section{Methodology for Relevance}
In this section, we introduce the methodology for the relevance determination module in Fig. \ref{fig: framework}.

% \subsection{Scene Processing}

% Let \(\mathcal{S}\) represent a scene consisting of a set of elements \(\mathcal{E} = \{e_1, e_2, \dots, e_n\}\), where $n$ is the number of elements in the scene, which is dynamic. Each element \(e_i\) is characterized by a set of characteristics $h(e_i)$. The naive representation of the scene \(\mathcal{S}\) is denoted as $\mathcal{S}_n$ represented by the tuple:
% \[
% \mathcal{S}_n = \left(\mathcal{E}, \mathcal{H} \right)
% \]
% where \(\mathcal{H} = \{h(e_1), h(e_2), \dots, h(e_n)\}\). $\mathcal{S}$ can be processed with perception algorithms, while $\mathcal{E}$ and $\mathcal{H}$ can be obtained at the same time. 

\subsection{Class and Element Scene Representation}
To quantify relevance efficiently and systematically, we propose a new representation $\mathcal{S}$. In $\mathcal{S}$, the scene is represented as a set of classes of elements: 

\begin{equation}
    \mathcal{S} = \{ C_1, C_2 \dots, C_m  \}
\end{equation}

where \(C_i\) represents the \(i\)th class of the classification results of all elements in \(\mathcal{E}\). $\mathcal{E}$ is the set of all elements in the scene. \(m\) is the number of classes. 

Each $C_i$  contains a set of elements classified into the class \(C_i\) according to the criterion $k_i \in k$, where $k=\{k_1, k_2, \dots,k_m\}$ is the set of classification criteria for each class. Thus, the set of elements belonging to the class \(C_i\) based on criterion \(k_i\) can be mathematically represented as:

\[
C_i = \{ e_j \mid e_j \in \mathcal{E}, e_j \text{ satisfies criterion } k_i \}.
\]

Each element \(e_j\) is included in the class \(C_i\) if it satisfies the criteria \(k_i\). 

In organizing elements, we consider two primary types of classes in this work: attribute classes and functional classes. Attribute classes group elements based on shared characteristics or properties, such as similar semantics or attributes. For example, attribute-based classes include general groupings like a class of fruits or books. Other examples within attribute classes are hierarchical classes, temporal classes, spatial classes, quantitative classes, etc.
Functional classes, on the other hand, group elements based on their roles, actions, or contributions to a common objective. Examples include behavioral classes, conceptual classes, and thematic classes. 

\subsection{Relevance Determination Mechanism}

% The goals of the relevance determination are three-fold. First, the relevance of each class $\mathcal{R}(\mathcal{C}_i)$ and the relevance of each element within relevant classes $\mathcal{R}(e_j)$ are quantified. Second, the set of relevant classes $\mathcal{C}_r$ and the set of relevant elements $\mathcal{E}_r$ are determined. Third, the information sufficiency for relevance quantification and determination is evaluated. The algorithm will ask for or wait for more information from the perception module to continue the relevance determination.  

We developed and proposed a multi-level relevance determination mechanism that works on the class level first and then on the element level within the relevant classes later. The relevance mechanism in both class and element level is defined as $\mathcal{M}$. The mechanism $\mathcal{M}$ takes the set of classes or elements and the set of features $\mathcal{F}$ as inputs. $\mathcal{F}$ can be represented as $\mathcal{F} = \{\mathcal{F}_v, \mathcal{F}_a, \mathcal{F}_c, \dots)$, where $\mathcal{F}_v$, $\mathcal{F}_a$, and $\mathcal{F}_c$ represent visual (e.g., elements detected, classes classified, motion information, etc.), auditory (e.g. conversation in the scene, etc.), and contextual (e.g., objective, tasks, etc.) features, respectively. %The perception module closely and actively monitors these features. 

There are two types of outcomes from the mechanism $\mathcal{M}$. Suppose the information is sufficient for the relevance quantification and determination. In this case, the outcomes are the relevance scores for each class or element, and, at the same time, the set of relevant classes $\mathcal{C}_r$ or the set of relevant elements $\mathcal{E}_r$. If the information in $\mathcal{F}$ is not sufficient for relevance quantification and determination, then the algorithm will refer to the perception module to request or obtain new cues $\mathcal{F}_{new}$ from the scene before continuing the processing. Thus, the mechanism $\mathcal{M}$ can be mathematically represented as:

% \mathcal{M}(\mathcal{F},\mathcal{X})=

\begin{equation}
   \mathcal{M}:\mathcal{F},\mathcal{X}\rightarrow\left\{
\begin{array}{ll}
\mathcal{R}(x) \text{ } \forall x\in\mathcal{X}\text{, and }\mathcal{X}_r & \text{if } \text{sufficient } \mathcal{F}\\
\text{ask or wait for } \mathcal{F}_{new}, & \text{else} 
\end{array} \right.
\end{equation}

where \( \mathcal{X} \) represents a set of classes or elements, \( x \) represents a specific class or element, and \( \mathcal{X}_r \) represents a set of relevant classes or elements.

%where \( \mathcal{X} \) represents the set of inputs to $\mathcal{M}$, \( x \) represents a specific input, and \( \mathcal{X}_r \) represents a set of relevant inputs. More specifically, when determining relevant classes, $\mathcal{X}$, $x$, and $\mathcal{X}_r$ refers to $\mathcal{S}$, $\mathcal{C}_i$, and $\mathcal{C}_r$, respectively. For relevant element determination,  $\mathcal{X}$, $x$, and $\mathcal{X}_r$ refers to $\mathcal{C}_r$, $e_j$, and $\mathcal{E}_r$, respectively. 

Both on the class level and the element level, the relevance $\mathcal{R}(x)$ is defined as the probability of $x$ being relevant given the objective of the human $\mathcal{O}$:

\begin{equation}
   \mathcal{R}(x) = P(x\mid\mathcal{O})
\label{eq: define r}
\end{equation}

(\ref{eq: define r}) can also be conditioned on other factors according to the requirements of the applications. After $\mathcal{R}(x)$ is quantified, there are two possibilities: $\mathcal{R}(x)$ is larger than a threshold $\tau_x$ implies that $x$ is relevant in the scene to the objective, and then $x$ is added to the relevant set $\mathcal{X}_r$:

\begin{equation}
R(x) > \tau_{\text{x}} \rightarrow x \text{ is relevant} \rightarrow \mathcal{X}_r = \mathcal{X}_r \cup x
\end{equation}

If $\mathcal{R}(x)$ is smaller than $\tau_x$, then $x$ is deemed as irrelevant:

\begin{equation}
R(x) < \tau_{\text{x}} \rightarrow  x \text{ is irrelevant}
\end{equation}

% With this methodology, the mechanism $\mathcal{M}$ can generate the desired outputs. 

(\ref{eq: define r}) can be influenced by a comprehensive set of possible factors. %, which can also categorized into Attribute Factors and Functional Factors. Attribute factors include spatial (e.g., proximity, accessibility, placement), temporal (e.g., urgency, task dependency, phase, step), and quantitative (e.g., quantity, weight, length, temperature, humidity) factors, etc. Functional factors include behavioral (e.g., necessity, suitability, efficiency, safety), human (e.g., capability, cognition, preference, culture.) factors, etc. 
In this paper, we compute (\ref{eq: define r}) based on four factors: the human's objective, tasks, preferences, and spatial placement.

\subsection{Class Relevance Modeling}
In this part, we model the class relevance based on (\ref{eq: define r}), which is defined as:

\begin{equation}
   \mathcal{R}(C_i) = P(C_i\mid\mathcal{O})
\label{eq: class relevance model}
\end{equation}

There could be several possible tasks $\mathcal{T}$ associated with the objective $\mathcal{O}$. Thus, we use the total probability to decompose (\ref{eq: class relevance model}) and obtain:

\begin{equation}
   \mathcal{R}(C_i) = \sum_{\mathcal{T}} P(C_i\mid\mathcal{T},\mathcal{O}) P(\mathcal{T}\mid \mathcal{O})
\label{eq: class relevance model decompose}
\end{equation}

In (\ref{eq: class relevance model decompose}), the first term $P(\mathcal{C}_i\mid \mathcal{T},\mathcal{O})$ represents the probability that class $C_i$ is relevant given the task is $\mathcal{T}$, and the objective is $\mathcal{O}$. The second term $P(\mathcal{T}\mid \mathcal{O})$ represents the probability that the current task is $\mathcal{T}$ given the objective is $\mathcal{O}$.
If the human's historical preference data about the specific class $C_i$ for the task $\mathcal{T}$ and objective $\mathcal{O}$ is available, then the first term can be derived as
% \begin{equation}
%    P(C_i\mid \mathcal{T},\mathcal{O}) = \mathscr{P}(C_i \mid \mathcal{T},\mathcal{O})
% \label{eq: class p prefernece }
% \end{equation}
%where $\mathscr{P}$ represents 
the probability of $C_i$ is relevant based on the preference. If the historical preference data is not available for class $C_i$ for task $\mathcal{T}$, the probability $P(C_i\mid\mathcal{T},\mathcal{O})$ can be derived from predictions based on the tools in the AI toolkit, such as Large Language Models (LLM), other datasets that incorporate action sequences, other human cues, etc. The second term in (\ref{eq: class relevance model decompose}) can be derived from prediction models for human tasks. 

If any term in (\ref{eq: class relevance model decompose}) is not available, or available, but with high uncertainty, the information to determine the relevant class is deemed as not sufficient, and $\mathcal{F}_{new}$ is required for accurate quantification of the class relevance. The uncertainty of the term $P(\mathcal{T}|\mathcal{O})$ can be estimated using entropy as: 

\begin{equation}
   H(P(\mathcal{T}|\mathcal{O})) = -\sum_{\mathcal{T}} P(\mathcal{T}|\mathcal{O})\ln{P(\mathcal{T}|\mathcal{O})}
\label{eq: task entropy}
\end{equation}

where $H$ represents the entropy of a probability. %The minimum value of (\ref{eq: task entropy}) is 0, corresponding to a certain prediction. The maximum value of (\ref{eq: task entropy}) is $\ln{n}$, where n is the number of possible outcomes of $P(\mathcal{T}|\mathcal{O})$. The maximum entropy corresponds to uniform probability over all possibilities, which means the prediction is uncertain. 

\subsection{Element Relevance Modeling}

With a similar methodology to class relevance modeling, the relevance for an element can be defined and modeled as: 

\begin{equation}
\begin{aligned}
   \mathcal{R}(e_j) &= P(e_j|\mathcal{O}) \\
   &= P(e_j|C_i, \mathcal{O})P(C_i|\mathcal{O}) \\
   &\phantom{=} +P(e_j|\neg C_i,\mathcal{O})P(\neg C_i|\mathcal{O})
\label{eq: element p}
\end{aligned}
\end{equation}

where $e_j \in C_i \in C_r$, and the symbol \( \neg \) denotes logical negation. $P(e_j|\neg C_i,\mathcal{O})$ represents the probability that $e_j$ is relevant given the class that $e_j$ belongs to is not relevant and the objective is $\mathcal{O}$, which equals to 0. Thus, (\ref{eq: element p}) becomes:

\begin{equation}
\mathcal{R}(e_j) = P(e_j|C_i,\mathcal{O})\mathcal{R}(C_i)
\label{eq: element p C}
\end{equation}

One unique consideration when modeling element relevance, but not for class relevance, is that some elements may share the same attributes except for the spatial factors. %For example, in the class of milk, there may be a couple of whole milk, reduced fat milk, or creamer. 
Thus, we model our element relevance based on a two-hierarchy approach in which each element is characterized by top-level attributes, such as type, and low-level attributes, such as spatial or temporal factors. With this consideration, (\ref{eq: element p C}) can be computed as:

\begin{equation}
\begin{aligned}
   \mathcal{R}(e_j) &= \mathcal{R}(C_i)P(e_j|C_i, \mathcal{O}) \\
   & = \mathcal{R}(C_i)(P(e_j|h_j,C_i,\mathcal{O})P(h_j|C_i,\mathcal{O})\\
   &\phantom{=}+P(e_j|\neg h_j,C_i,\mathcal{O})P(\neg h_j|C_i,\mathcal{O}))\\
   &=R(C_i)P(e_j|h_j,C_i,\mathcal{O})P(h_j|C_i,\mathcal{O})
\label{eq: element final}
\end{aligned}
\end{equation}

where $h_j$ represents the event that the high-level attributes of $e_j$ are relevant. The value $P(e_j|\neg h_j,C_i,\mathcal{O})=0$ in the above equation. 

%To better interpret (\ref{eq: element final}), an explanation for each term is given. The first term $R(C_i)$ is simply the relevance for the class $C_i$, which $e_j$ is classified into. The second term means the probability of $e_j$ is relevant given the high-level attributes of $e_j$ are relevant, the objective is $\mathcal{O}$ and the class $C_i$ is relevant, which can be modeled based on low-level attributes, such as spatial or temporal factors. The third term represents the probability that the high-level attributes of $e_j$ are relevant given $C_i$ is relevant and the objective is $\mathcal{O}$, which can be derived based on preference or prediction, similar to the methods for class relevance modeling. 

When determining if the information is sufficient for relevant element determination, the availability and/or uncertainty of the three terms in (\ref{eq: element final}) should be carefully examined similarly to relevant class determination. 

\subsection{Constraints and Dependencies}

Constraints and dependencies among elements, such as temporal and spatial constraints, can be considered in a post hoc manner for the downstream decision-making modules after relevance is determined. If a relevant element depends on or is constrained by an element, this element should also be considered in relevance and decision-making. It is critical to consider these constraints so that the downstream decision-making module can be informed comprehensively and produce better actions.

\section{Simulation Evaluation}
A comprehensive simulation evaluation is conducted in this paper for the aspects of relevance quantification and its application to HRC.

\subsection{Simulation Domain Setup}
We introduced two distinct HRC testing domains for our evaluations with detailed description as follows: 

\textbf{Coffee}: %The Coffee domain involves making coffee with creamer and passing it to the human. The original simple problem formulation comprises 19 objects without task dependencies. In the randomly generated problem instances, 10 to 30 objects are randomly selected from a pool of 200 task-irrelevant kitchenware items and incorporated into the environment. Additionally, three random spatial constraints are introduced, creating task dependencies. In contrast, the hard problem formulation originally includes 36 objects, with randomly generated instances adding 20 to 50 objects. The number of spatial constraints increases to eight in the hard formulation. Across both cases, the spatial constraints involve one object being positioned on top of another, requiring the top object to be moved before the lower object can be accessed. 
The Coffee domain involves making coffee with creamer and passing it to humans. The original simple problem setup includes 19 objects with no task dependencies. In randomly generated cases, 10 to 30 elements from 200 irrelevant kitchenware items are added, along with three random spatial constraints introducing task dependencies. The hard problem setup starts with 36 elements, and randomly generated cases add 20 to 50 more, with eight spatial constraints. In both cases, the constraints require moving an object on top of another to access the lower one.

\textbf{Cereal}: %The Cereal domain involves the robot arm preparing cereal using all necessary objects, delivering it to the human, and returning all used items to their original locations. The original simple problem formulation includes 18 objects and naturally incorporates 20 task dependencies due to the storage locations and the presence of containers. For instance, the cereal is stored in a cabinet, which must be opened before the cereal can be accessed, and the cereal box itself needs to be opened before it can be poured into a bowl. Random problem instances are generated by introducing 10 to 45 additional task-irrelevant kitchenware items into the environment. The hard problem formulation initially consists of 26 objects with 20 task dependencies, along with an additional 30 to 60 objects introduced for the randomly generated instances. Note that no additional task dependencies are introduced in these randomly generated problems.
The Cereal domain focuses on a robot preparing cereal, serving it to a person, and returning the items to their original places. In the original simple problem, there are 18 elements and 20 natural task dependencies, such as needing to open a cabinet to access the cereal. In random problem instances, 10 to 45 irrelevant kitchenware items are added. The hard version starts with 26 objects and 20 task dependencies, with an additional 30 to 60 irrelevant objects added in random cases without new task dependencies.

\subsection{Relevance Quantification Evaluation}

For the relevance quantification evaluation, we adopted and developed several metrics. First, we define precision $\mathfrak{P}$, recall $\mathfrak{R}$, and $F_1$ score $\mathfrak{F}$, which evaluate the prediction of relevant element set based on the ground truth of the relevant element set. %Precision $\mathfrak{P}$ assesses the proportion of predicted elements that are genuinely relevant to the task. A higher threshold for the class and/or elements will increase $\mathfrak{P}$. Recall is particularly crucial since missing an element in the relevance prediction could cause downstream tasks to fail. $F_1$ score is to balance the precision and recall. 
%which represents the precision of the prediction evaluated on the ground truth of the relevant element set. This metric assesses the proportion of predicted elements that are genuinely relevant to the task. A higher threshold for the class and/or elements will increase $\mathfrak{P}$. Second, we defined the recall metric, i.e., $\mathfrak{R}$, which measures the recall rate of the prediction based on the ground truth of the relevant element set. Recall is particularly crucial since missing an element in the relevance prediction could cause downstream tasks to fail. Third, to balance the accuracy and recall, we introduced $\mathfrak{F}$, representing the $F_1$ scores of the prediction, which is defined as $\mathfrak{F} = 2 \cdot \mathfrak{P}\cdot\mathfrak{R} /(\mathfrak{P}+\mathfrak{R})$. 
Moreover, we introduced a metric, $\mathfrak{N}$, representing the number ratio of predicted relevant elements to actual relevant elements for assessing speed improvement in downstream tasks. A higher $\mathfrak{N}$ will probably slow down the downstream tasks. Those metrics systematically evaluate the relevance quantification from aspects of effectiveness, completeness, and efficiency. Different applications will require different combinations of optimal metrics.

In this paper, we emphasize the simulation for the coffee domain for the brevity and clarity of our analysis. In the simulation, the objective is predefined as ``get something to drink at the break of a conference", and the task is defined as ``drink cold brew coffee". A large language model (LLM) GPT-4o is adopted to classify all the elements into the scene representation $\mathcal{S}$ and provide the probabilities necessary for the relevance quantification, as illustrated in Section IV. We tested 25 combinations of the class thresholds $\tau_c$ and element thresholds $\tau_e$, with 30 simple cases randomly generated for each combination.  

\begin{figure*}[!t]
\begin{center}
\includegraphics[width=\textwidth]{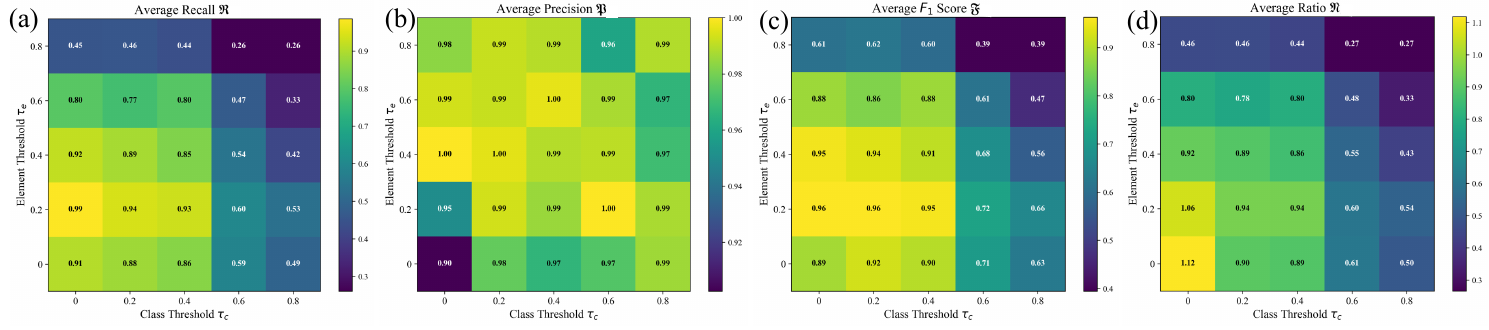}
\end{center}
\caption{Simulation results for the coffee domain. % with (a) recall based on relevant element ground truth  $\mathfrak{R}_r$; (b) precision based on relevant element ground truth $\mathfrak{P}_r$; (c) $\mathfrak{F}_r$ based on relevant element ground truth; (d) ratio between numbers of predicted relevant elements and ground truth relevant elements $\mathfrak{N}_r$; (e) recall based on necessary element ground truth $\mathfrak{R}_n$; (f) precision based on necessary element ground truth $\mathfrak{P}_n$; (g) $\mathfrak{F}_n$ based on necessary element ground truth; (h) ratio between numbers of predicted relevant elements and ground truth necessary elements $\mathfrak{N}_n$. 
 The values in the figures are averaged across 30 cases for each threshold combination. When $\tau_c$ and $\tau_e$ equal to 0.2, our methodology archives a recall $\mathfrak{R}$ of 0.94, a precision $\mathfrak{P}$ of 0.99, an F1 score $\mathfrak{F}$ of 0.96, and an object ratio $\mathfrak{N}$ of 0.94.} 
\label{fig: sim results}
\end{figure*}

The simulation results are shown in Fig. \ref{fig: sim results}, which agree well with common sense. As $\tau_c$ and $\tau_e$ increase, more classes and elements will be pruned. Thus, the recall decreases, the precision increases, and the relevant object ratio decreases as these two thresholds increase. The F1 scores demonstrate a trend that increases first and then decreases with the threshold increase. %When we compare the metrics for the ground truth of relevant elements and necessary elements, $\mathfrak{R}_n$ is higher than $\mathfrak{R}_r$, $\mathfrak{P}_n$ is smaller than $\mathfrak{P}_r$, and $\mathfrak{N}_n$ is larger than $\mathfrak{N}_r$. This is because the set of necessary elements is a subset of the set of relevant elements, 

The trend of $\mathfrak{R}$ in Fig. \ref{fig: sim results}(a) demonstrates that when $\tau_c$ and $\tau_e$ equal to or are smaller than 0.2, most relevant elements can be retained. At the same range of the thresholds, according to Fig. \ref{fig: sim results}(d), the number of relevant elements predicted is very close to the number of relevant elements in the ground truth, which is very small when compared with the total number of elements in the scene. This demonstrates that our relevance quantification methodology can successfully remove most irrelevant elements while preserving relevant elements.

\begin{figure}[!t]
\begin{center}
\includegraphics[width=0.8\columnwidth]{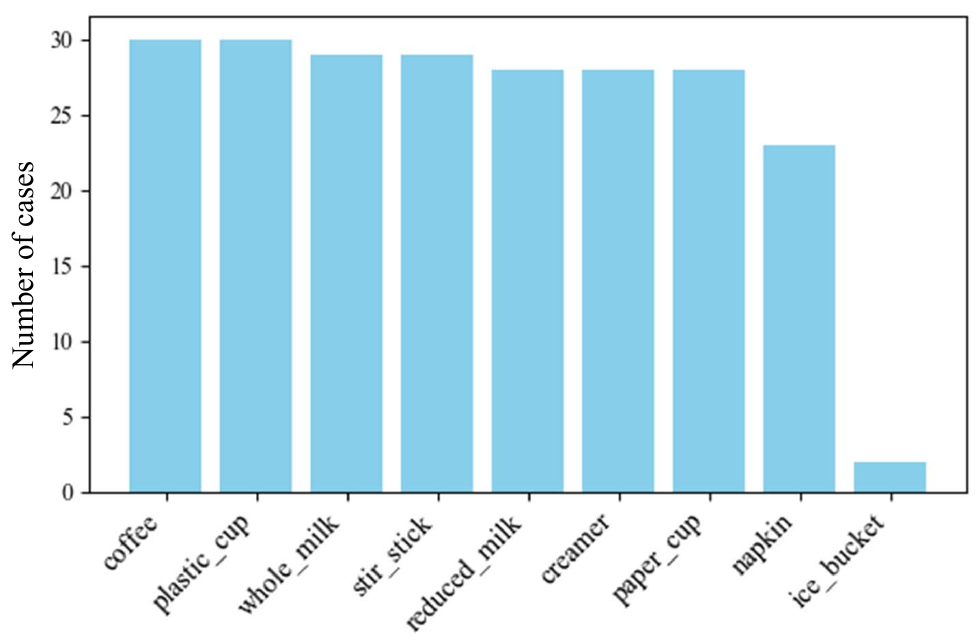}
\end{center}
\caption{Number of cases each element is predicted to be relevant out of 30 cases without considering constraints. $\tau_c$ and $\tau_e$ are 0.2. Our relevance quantification method predicts relevance accurately and reliably. } 
\label{fig: element historgram}
\end{figure}

We select the thresholds $\tau_c$ and $\tau_e$ to be 0.2. At this threshold combination, our methodology archives a recall $\mathfrak{R}$ of 0.94, a precision $\mathfrak{P}$ of 0.99, an F1 score $\mathfrak{F}$ of 0.96, and an object ratio $\mathfrak{N}$ of 0.94. These results demonstrate that our methodology is effective and accurate in dimension reduction. The number of cases in which each type of element is relevant is shown in Fig. \ref{fig: element historgram} without considering the constraints. Of the 30 cases, a couple of them randomly add an ice bucket to the setup and predict that the ice bucket is relevant, which agrees with common sense in the context of drinking cold brew coffee. All 30 cases predict that coffee and plastic cups are relevant. These results further demonstrate that relevance prediction is very accurate.

\subsection{Relevance in HRC}
One benefit of relevance in HRC is that relevance can help the robot better reason about the scene and the human's requirements in a well-structured manner, which resembles how humans interact with other humans. This benefit can be demonstrated with a simple robot inquiry example in a coffee serving setup. The human-like robot will first detect the relevant classes and types of elements. If the relevance of any classes or types of elements is high, those classes or types are deemed necessary (coffee in this domain). Other relevant classes or types of elements are optional, and the robot asks the human if they need those relevant elements. The number of inquiries required using different methods is shown in Table \ref{table: inquiry}. Note that the number of inquiries required without relevance equals the average number of objects in the testing setup. It is shown that with accurate and proper reasoning about the relevance of an element to a task or objective, the number of inquiries required for appropriate assistance is reduced by 80.84\% compared with the number without relevance. Thus, human-robot interactions can be much more natural and fluent.

\begin{table}[t]
\caption{Number of inquiries required for appropriate assistance. Relevance dramatically reduces the number of inquiries.}
\begin{tabular}{ccc}
\hline
method                                                                       & inquiry elements                                                                                                              & count      \\ \hline
\textbf{\begin{tabular}[c]{@{}c@{}}relevance \\ with necessity\end{tabular}} & \textbf{\begin{tabular}[c]{@{}c@{}}creamer, plastic cup, paper cup, \\ whole milk, reduced milk, stick, napkin\end{tabular}} & \textbf{7} \\
relevance                                                                    & \begin{tabular}[c]{@{}c@{}}coffee, creamer, plastic cup, paper cup, \\ whole milk, reduced milk, stick, napkin\end{tabular}  & 8          \\
no relevance                                                                 & Everything in the scene                                                                                                      & 36.53      \\ \hline
\end{tabular}
\label{table: inquiry}
\end{table}

\subsection{Relevance in Task Planning}
This section demonstrates the effectiveness of relevance in task planning. A Fast Downward planner is employed. We compare three methods: our relevance with 0.2 as thresholds, random relevance with 0.2 as thresholds, and pure planning without relevance. For each method, we evaluate the processing time (relevance computation + planning), the timeout rate (120s), and the failure rate. Results are shown in Table \ref{tab:comparison}. Note that the failure rate is not applicable to pure planning, as solutions are guaranteed to be found if they exist.

Relevance dramatically improves task planning performance. Compared with Pure Planning, relevance significantly reduces the time cost by up to 79.56\% and the timeout rate across all problem formulations. As the problem formulation becomes more complex, we observe a notable increase in the timeout rate and planning time for Pure Plan, highlighting its inefficiency in solving complex task-planning problems. When compared with random relevance, we observe a substantial reduction in failure rates, demonstrating that our relevance successfully and accurately determines the relevant elements of the scene. Moreover, the highest failure rate after adopting relevance in the four test cases is 0.04, which demonstrates that our relevance can effectively, accurately, and robustly reduce the task planning time.

\begin{table}[]
\centering
\caption{Task Planning Performance Comparison. Our methodology of relevance determination accurately predicts relevance and dramatically reduces the planning time.}
\begin{tabular}{ccccc}
\hline
Domains                        & Metrics & Relevance      & \begin{tabular}[c]{@{}c@{}}Pure \\ Planning\end{tabular} & \begin{tabular}[c]{@{}c@{}}Random \\ Relevance\end{tabular} \\ \hline
\multirow{3}{*}{Coffee simple} & Time    & \textbf{16.76} & 45.84                                                    & 21.61                                                       \\
                               & Timeout & \textbf{0.00}  & 0.22                                                     & 0.05                                                        \\
                               & Fail    & 0.04           & -                                           & 0.62                                                        \\ \hline
\multirow{3}{*}{Coffee hard}   & Time    & 36.91          & -                                                        & \textbf{21.38}                                              \\
                               & Timeout & \textbf{0.00}  & 1.00                                                     & 0.02                                                        \\
                               & Fail    & 0.01           & \textbf{0.00}                                            & 0.88                                                        \\ \hline
\multirow{3}{*}{Cereal simple} & Time    & \textbf{13.37} & 63.84                                                    & 39.61                                                       \\
                               & Timeout & \textbf{0.00}  & 0.38                                                     & 0.05                                                        \\
                               & Fail    & 0.02           & -                                            & 0.58                                                        \\ \hline
\multirow{3}{*}{Cereal hard}   & Time    & \textbf{30.70} & -                                                        & 53.82                                                       \\
                               & Timeout & \textbf{0.00}  & 1.00                                                     & 0.12                                                        \\
                               & Fail    & 0.04           & -                                            & 0.60                                                        \\ \hline
\end{tabular}
\label{tab:comparison}
\end{table}

\subsection{Relevance in Perception}
Our previous work demonstrated that by selectively processing the regions containing relevant elements, a notable maximum reduction of 30.09\% in inference time and 26.53\% in total time per frame of an object detector is achieved. Additionally, relevance improves two safety metrics by 11.25\% and 13.50\%, respectively \cite{zhang2024does}.

\section{Experimental Demonstration}
In this section, we present the real-world demonstration to verify the effectiveness of our proposed framework and relevance determination methodology. 

\subsection{Experimental Setup}
The experimental setup is shown in Fig. \ref{fig: demo}(a). On one end of the table, a UR5 robot arm with a robotiq gripper is mounted to reason about the relevance and generate actions to assist humans. The table is a snack table for a conference with desserts, drinks, and utensils. A microphone is placed on the table to pick up the audio information and cues in the scene. The HRC task is for the robot to assist two humans into the scene one by one. The robot utilized relevance for optimal decision-making and HRC. 

% \begin{figure}[!t]
% \begin{center}
% \includegraphics[width=\columnwidth]{setup_1.jpg}
% \end{center}
% \caption{Experimental setup. } 
% \label{fig: experimental setup}
% \end{figure}

The demonstration is implemented in Python using the \texttt{threading} package for multi-threading and asynchronous computation. The communication between threads is achieved using \texttt{Event} and \texttt{Queue}. At the beginning of the code, we initialized two threads, as shown in the perception module of Fig. \ref{fig: framework}, one for the visual information processing using the OpenAI API \cite{GPT4O} and another one for picking up the microphone and processing the audio information using Assembly AI API \cite{assemblyai2025}. In the visual information processing thread, we currently implement a VLM model for visual information processing, extracting the semantic information of all the objects on the table, classifying the objects into class and element representation, and detecting the human motion and objectives. For each iteration in the visual threads, the trigger criteria are checked. We implement the trigger criterion, the changes in human numbers in the scene, which is sufficient for our current demonstration. Once a trigger criterion is met, a new thread for relevance determination is initialized, and relevance is determined based on the methodologies in Section IV. The probabilities required for the computation are derived from an LLM model in the AI toolkit. We currently leverage the OpenAI API, and the model we used is GPT-4o \cite{GPT4O}. Within the same thread, the action is generated based on the relevance identified to best assist humans.

\subsection{Demonstration Results}
\begin{figure}[!t]
\begin{center}
\includegraphics[width=\columnwidth]{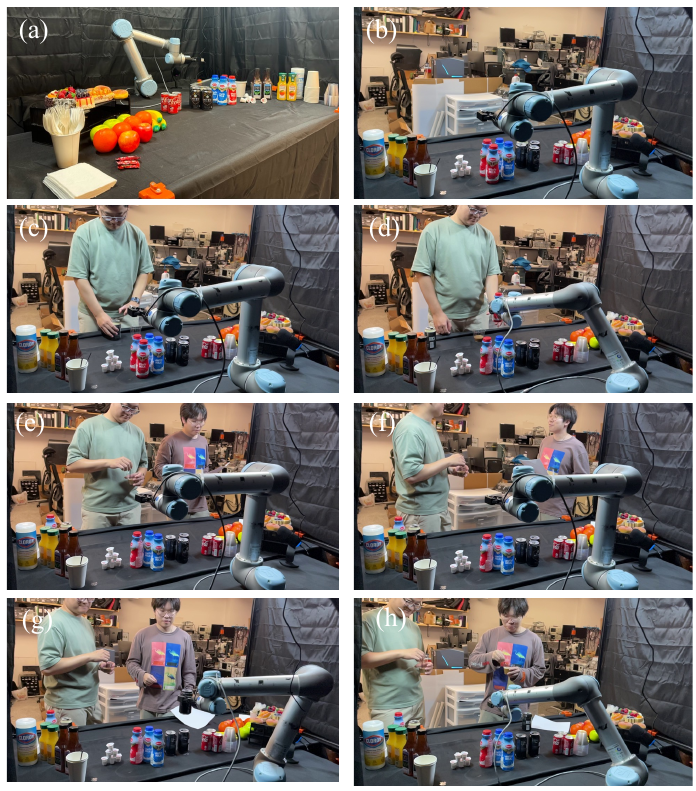}
\end{center}
\caption{The illustration of (a) experimental setup and (b-h) demonstration results. With relevance, the robot successfully and seamlessly assists two human agents with cold-brew coffee drinking.} 
\label{fig: demo}
\end{figure}

The demonstration results are shown in Fig. \ref{fig: demo}(b-h). In Fig. \ref{fig: demo} (b), in the absence of a human in the scene, due to the lack of triggers, the perception module is continuously running, and no relevance determination is initialized. In Fig. \ref{fig: demo}(c-d), a human agent with recorded preferences for coffee enters the scene and grabs a coffee. The trigger is activated, and the human's objective and relevance are determined based on the methodologies we described in Section IV. Based on the recorded preferences, the robot actions are automatically generated to serve the human agent whole milk and a stir stick. In Fig. \ref{fig: demo}(e), a second human enters without indicative actions. The trigger is activated, but the relevance determination module lacks sufficient cues. The system continues looping through the perception module to collect more information. In Fig. \ref{fig: demo}(f-g), a conversation between two human agents about coffee provides more cues to determine relevance. Without preference for the second human, the robot first generates actions related to the necessary elements (i.e., the coffee) and then inquires about the second human’s preferences for making the coffee. Based on the human response, the system generates a complete action sequence to serve the second human agent. In Fig. \ref{fig: demo}(h), two humans successfully get their preferred coffee with the robot's assistance, demonstrating the effectiveness of relevance. Without the unique components in relevance, such as the evaluation of cue sufficiency, a flexible probabilistic framework integrating all the factors, a novel hierarchical scene representation, etc., this type of HRC assistance in a general setup will be challenging. The demonstration video can be found here: https://www.youtube.com/watch?v=XVUxWbwsYhs

\section{Conclusion}
This paper presents a novel concept, dimensionality reduction approach, and HRC framework termed `relevance'. Relevance is a dimensionality reduction process that incorporates a continuously operating perception module, evaluates cue sufficiency within the scene, and applies a flexible formulation and computation framework considering diversified factors. The perception module runs continuously and leverages multiple modalities of sensors and processing to provide cues and information. The event-based framework has the capability to determine if the relevance determination should be triggered. A probabilistic methodology to evaluate relevance based on a novel hierarchical scene representation can efficiently and accurately quantify relevance, considering various human and environmental factors. Simulations demonstrate the framework’s effectiveness, achieving a precision of 0.99, a recall of 0.94, an F1 score of 0.96, and an object ratio of 0.94.
%. We demonstrate that our methodology can effectively and accurately determine relevance. The experimental results underscore the importance of relevance to enhance natural human-robot interaction, improve perception performance, and optimize task planning and decision-making in terms of both time and accuracy. We consider our work as a step toward developing a general-purpose robotic assistant.
%One limitation we identified is the reliance on large language models (LLMs) to assign probabilities for determining the relevance of a class, given the task and objective in the absence of preferences. LLM introduces the major time costs in our framework and can occasionally result in the omission of relevant classes and elements in a scene. Future work could explore the integration of semantic relevance and human intention prediction into the relevance formulation. Another area for future research involves optimizing the framework's architecture by allowing modules to run at different frequencies based on task demands, likely shifting the current serial framework to a more parallelizable architecture. Additionally, a more systematic and general approach to handling spatiotemporal constraints could be developed to address a wider range of constrained task-planning problems.
Relevance can significantly enhance HRC by accurately and effectively improving task planning time by 79.56\% compared to pure planning for a cereal task, reducing perception latency by up to 26.53\% for object detection, enhancing safety by up to 13.50\%, and decreasing the number of inquiries by 80.84\%. The experimental demonstration of serving two people coffee drinking shows that relevance enables the robots to assist humans in a proactive, seamless, and structured manner. This is a significant advancement toward creating a versatile robotic assistant capable of providing intelligent support to humans.

\addtolength{\textheight}{0cm}   % This command serves to balance the column lengths

%%%%%%%%%%%%%%%%%%%%%%%%%%%%%%%%%%%%%%%%%%%%%%%%%%%%%%%%%%%%%%%%%%%%%%%%%%%%%%%%

%%%%%%%%%%%%%%%%%%%%%%%%%%%%%%%%%%%%%%%%%%%%%%%%%%%%%%%%%%%%%%%%%%%%%%%%%%%%%%%%

%%%%%%%%%%%%%%%%%%%%%%%%%%%%%%%%%%%%%%%%%%%%%%%%%%%%%%%%%%%%%%%%%%%%%%%%%%%%%%%%

%\newpage

\bibliographystyle{IEEEtran}
\bibliography{IEEEabrv,references}

\end{document}